\documentclass[11pt]{article}
\usepackage{acl}

\usepackage{times}
\usepackage{latexsym}
\usepackage[T1]{fontenc}
\usepackage[utf8]{inputenc}
\usepackage[expansion=false]{microtype}
\IfFileExists{inconsolata.sty}{\usepackage{inconsolata}}{}
\usepackage{graphicx}
\usepackage{booktabs}
\usepackage{amsmath}
\usepackage{amssymb}
\usepackage{multirow}
\usepackage{xcolor}
\usepackage{colortbl}
\usepackage{enumitem}

\definecolor{tablepos}{HTML}{2A9D8F}
\definecolor{tableneg}{HTML}{D1495B}
\definecolor{tablepostext}{HTML}{006D62}
\definecolor{tablenegtext}{HTML}{9B1D32}
\newcommand{\tabpos}[2]{\cellcolor{tablepos!#1}\ensuremath{+#2}}
\newcommand{\tabneg}[2]{\cellcolor{tableneg!#1}\ensuremath{-#2}}

\title{Empathy Is Steerable but Multi-Axial: Mechanism Geometry and Persona Effects in LLMs}

\author{JuHeon Ha \quad Byounghan Lee \quad Yunseo Choi \quad
  Kyung-Ah Sohn\thanks{Corresponding author} \\
  Department of Artificial Intelligence, Ajou University \\
  \texttt{\{hjhj97, qudgks96, chldbstj0512, kasohn\}@ajou.ac.kr}}

\begin{document}
\maketitle

\begin{abstract}
  
  Activation steering has been used to control traits such as honesty, refusal, and sycophancy, yet supportive empathy is evaluated along multiple dimensions that need not correspond to independently controllable activation directions. 
  Using the EPITOME framework, which decomposes supportive empathy into Emotional Reactions, Interpretations, and Explorations, we study three instruction-tuned LLMs and ask whether candidate directions derived from these labels produce distinguishable intervention effects or instead share structure, and how persona prompts interact with those directions.  
  We find that contrastive activation addition yields a stable middle-layer intervention that consistently shifts the EPITOME proxy scores across models, moving empathy analysis beyond response-level scoring. However, the recovered directions are only partially separable: steering one direction induces off-target shifts, and hand-crafted prompting shifts the empathy profile rather than isolating a single dimension. 
  Persona prompts substantially change EPITOME scores, but a paired activation-shift decomposition shows the recovered subspace captures only about $3\%$ of persona-induced squared activation-shift magnitude at layer 15. Under this EPITOME-based definition, expressed empathy is steerable but multi-axial, and  controlling persona-conditioned empathy requires targeting structure beyond individual mechanism directions.\footnote{Code is available in our GitHub repository:\newline \href{https://github.com/hjhj97/Empathy-Is-Steerable-but-Multi-Axial}{\textcolor{darkblue}{github.com/hjhj97/Empathy-Is-Steerable-but-Multi-Axial}}}

\end{abstract}



\begin{figure}[t]
\centering
\includegraphics[width=0.9\columnwidth]{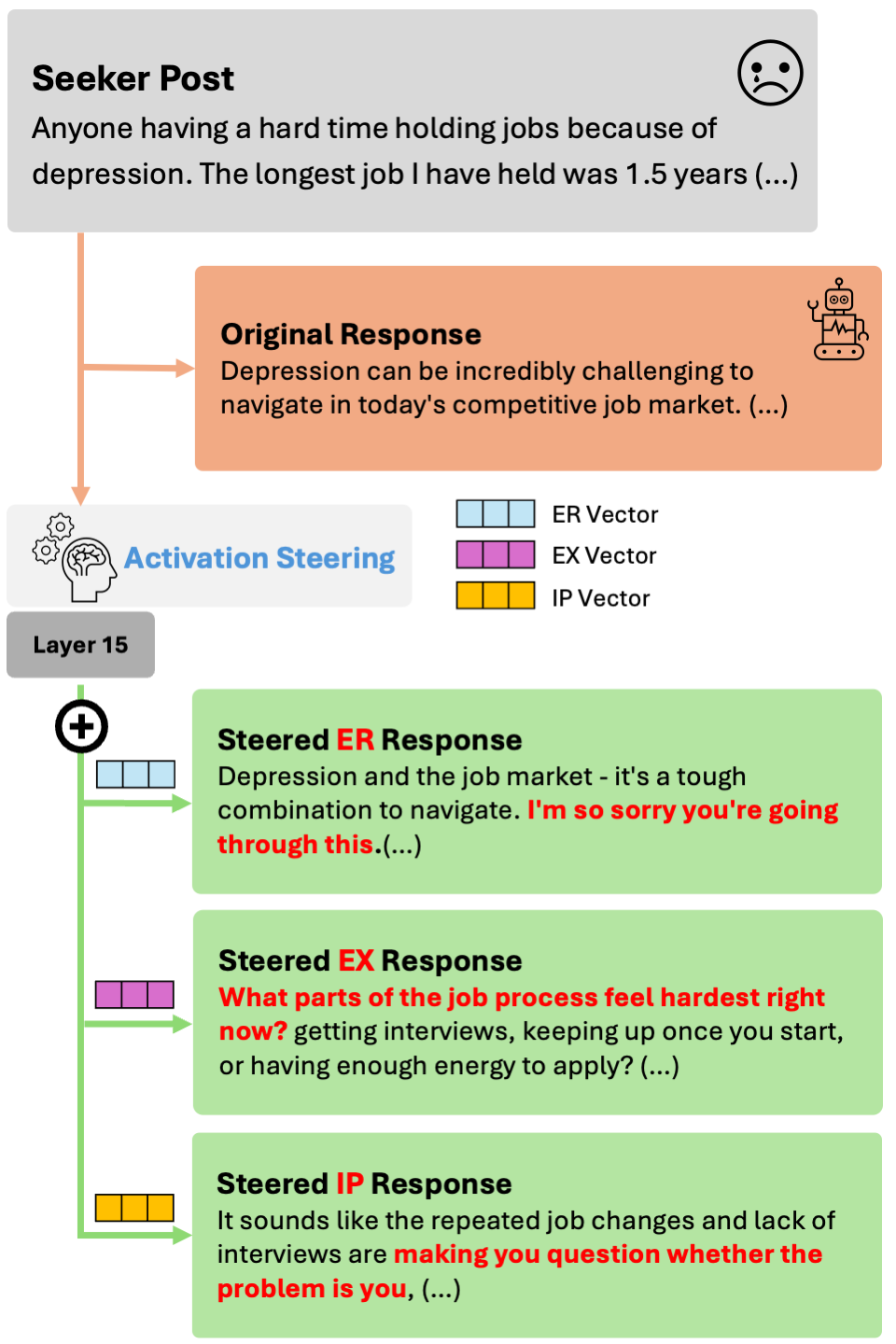}
\caption{Example of mechanism-level steering on a support-seeking post. Highlighted phrases mark the mechanism-characteristic content introduced by each steering vector.}
\label{fig:steering-example}
\end{figure}

\section{Introduction}


Conversational LLMs are evaluated as emotional-support systems, and their responses are rated as more empathic than human-written ones~\citep{sharma2020epitome,welivita2024perceived}. But a high empathy rating says little about \textit{how} a response is empathic. Empathy is not one thing: it includes affective resonance with what the seeker feels, cognitive understanding of their situation, and communicative moves that invite elaboration~\citep{sharma2020epitome,maurya2024multidim}. A response can realize one component while failing another, so an aggregate score reveals little about how a model is empathic. Fine-grained control matters because a system may need to strengthen one supportive behavior without amplifying another; activation steering provides an inference-time control channel without retraining or adding empathy instructions to the user-facing prompt.

Empathy is abstract and partly subjective, so analyzing it mechanistically requires decomposing broad empathy judgments into discrete, behaviorally anchored components.
In this work, we use \textit{empathy} to refer to EPITOME-defined supportive empathy in peer-support dialogue. EPITOME~\citep{sharma2020epitome} defines three communication mechanisms --- Emotional Reactions (ER), Interpretations (IP), and Explorations (EX) --- each scored on a 0/1/2 scale by a classifier trained on human-annotated support dialogues. ER captures emotional acknowledgment, IP captures inferred understanding of the seeker's situation, and EX captures questions or invitations to elaborate;
Figure~\ref{fig:steering-example} illustrates these categories: ER steering adds emotional acknowledgment, IP steering adds an interpretation, and EX steering adds a follow-up question. However, separating empathy into three evaluation categories does not imply that the model treats them independently.
Response-level scores show co-variation, not independent control; we therefore intervene along each candidate direction and measure target and non-target effects.

For other high-level traits, such an intervention is available. Activation steering localizes a behavior as a direction in the residual stream and adds or subtracts it at inference to control that behavior~\citep{panickssery2024steering,turner2023actadd}, and the approach has been validated for honesty, refusal, sycophancy, and demographic bias~\citep{chen2025persona,li2025fairsteer}. Here, steering is not merely a way to raise empathy scores, but a causal probe of representational organization. Prior work has studied generated responses under prompting, fine-tuning, and persona conditioning~\citep{welivita2024perceived,kim2025empathygaps,shen2025multidemo}, but not whether empathy is internally steerable as one quantity or as multiple interacting mechanisms. 


Persona makes this question concrete: persona prompts can make models express systematically different levels of empathy depending on the specified demographic~\citep{kim2025empathygaps,shen2025multidemo}. Whether a persona shifts empathy \textit{uniformly} or redistributes it across ER, IP, and EX determines whether a single steering direction can close persona-conditioned empathy gaps.

Within this framework, we first establish that contrastive activation addition yields reliable empathy-shifting directions with a consistent middle-layer intervention point across models. Building on this, we ask two questions. \textbf{(RQ1)} \textit{Do ER, IP, and EX occupy separable representational directions, or do they share subspace structure that limits independent control?} \textbf{(RQ2)} \textit{Do persona prompts shift empathy along the recovered mechanism directions, and how much of persona-induced empathy variation is captured by the recovered ER/IP/EX mechanism basis?}
Both questions require intervening on the representation rather than scoring outputs, so we extract a steering vector per mechanism and combine layer-sweep, geometric, and projection analyses across three open instruction-tuned models.



Our findings complicate the natural single-axis reading. \textbf{For RQ1}, the recovered ER/IP/EX directions are not independently separable but share representational structure and produce off-target shifts. \textbf{For RQ2}, persona prompts produce large empathy-score changes without comparable movement along the recovered mechanism directions; the ER/IP/EX subspace captures only about $3\%$ of persona-mean squared activation-shift magnitude at layer 15. We make three main contributions:




\begin{itemize}[leftmargin=*, noitemsep, topsep=0pt]
\item \textbf{A cross-model activation-level handle on empathy.} A layer sweep across three open LLMs identifies a consistent middle-layer steering point, robust under a second automated evaluator and showing partial out-of-domain transfer, moving empathy from response-level scoring to a tractable target for mechanistic analysis.


\item \textbf{Expressed empathy is multi-axial in LLM representations, not a single controllable trait.} Geometry analysis shows that the three mechanism directions share representational structure rather than forming independent axes, producing measurable behavioral spillovers across mechanisms that prompting alone cannot isolate. 


\item \textbf{Persona-induced empathy bias is largely outside the recovered mechanism basis.} Large changes in persona-conditioned empathy scores correspond to only small projection shifts along the recovered mechanism directions,
positioning single-vector debiasing as insufficient for closing persona-induced empathy gaps.
\end{itemize}

\begin{figure*}[t!]
\centering
\includegraphics[width=\textwidth]{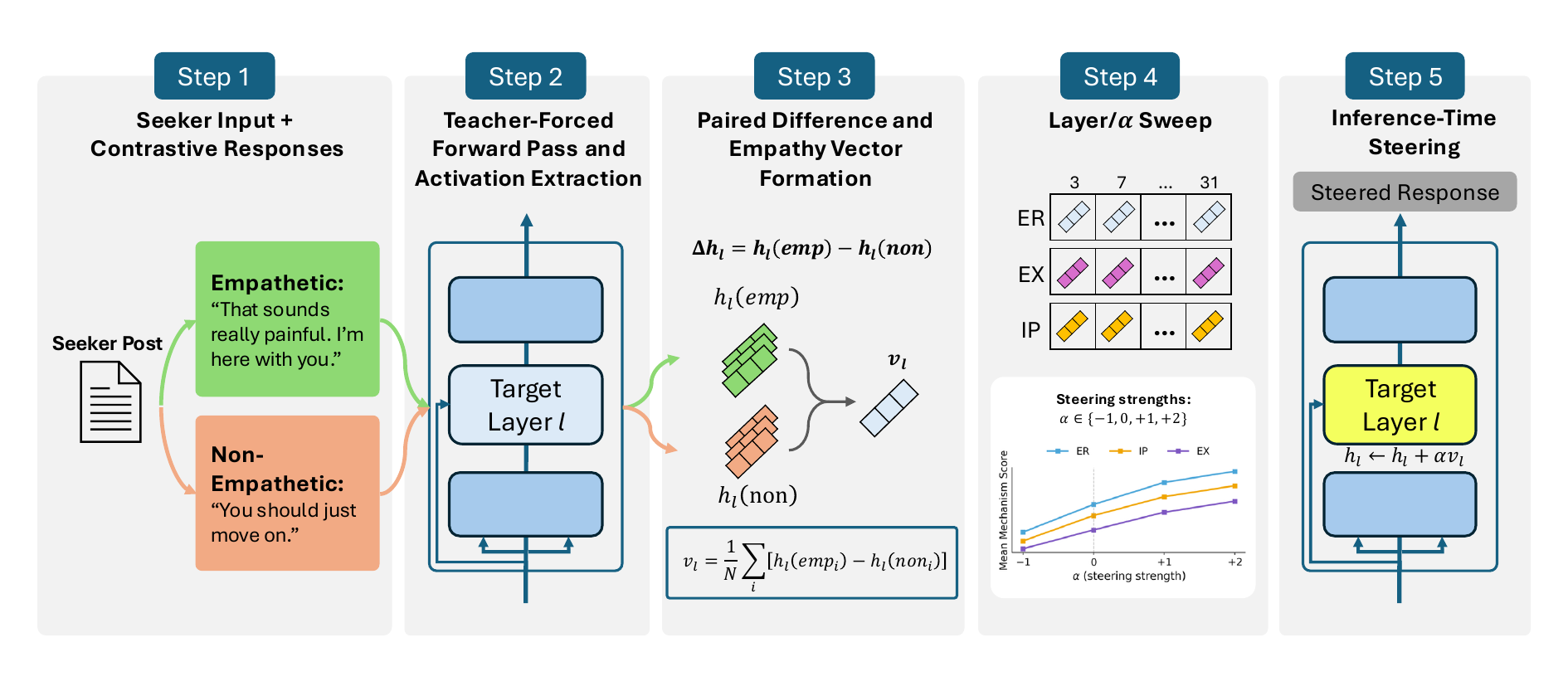}
\caption{%
Empathy vector extraction and inference-time steering. (1) \textbf{Contrastive examples.} For each EPITOME mechanism, we sample high- and low-empathy seeker--response rows from the filtered data. (2) \textbf{Teacher-forced activation extraction.} We run each seeker--response sequence through the model and read the layer-$\ell$ final-response-token residual state. (3) \textbf{Empathy-vector formation.} We form $v_\ell^m$ by subtracting the low-empathy pool mean from the high-empathy pool mean. (4) \textbf{Layer/$\alpha$ sweep.} We evaluate candidate layers and steering strengths to select a stable operating point. (5) \textbf{Inference-time steering.} During generation, a forward hook updates $h_{\ell,t} \leftarrow h_{\ell,t} + \alpha v_\ell^m$ to produce the steered response.%
}
\label{fig:main-overview}
\end{figure*}

\section{Related Work}
\label{sec:related}

\noindent\textbf{Activation steering.}
Activation steering modifies model behavior by adding fixed direction vectors to residual-stream activations at inference.
\citet{turner2023actadd} introduced ActAdd using contrastive prompt pairs, and \citet{panickssery2024steering} formalized Contrastive Activation Addition (CAA) with the last-token mean-difference recipe we follow, localizing the effect to middle layers.
Earlier representation-engineering work treats internal directions as analysis objects, not only control knobs~\citep{subramani2022steering,zou2023representation}. \citet{li2024iti} identify truthful directions inside attention heads, a head-level analogue to our residual-stream localization, and related work finds linear structure for refusal and truthfulness~\citep{arditi2024refusal,marks2023geometry,park2024linear}. \citet{wang2025logical} report that knowledge vectors for logical reasoning occupy independent subspaces, motivating our geometric test, which we partially reject for empathy.

\noindent\textbf{Trait and emotion vectors.}
\citet{chen2025persona} (Persona Vectors) extract per-trait directions for behaviors such as evil and sycophancy. Their setup is closest to ours, but treats each trait as a single independently controllable axis; we instead characterize geometric separability and shared structure among empathy mechanisms. \citet{feng2026persona} compose personality vectors via vector algebra, and \citet{sun2025personality} modulate Big-Five traits through model merging. \citet{ku2025valencearousal} decompose LLM emotion vectors into a valence--arousal subspace, supporting the view that affective representations are multidimensional --- consistent with our finding that ER, EX, and IP vectors share a partially overlapping subspace. Unlike these decomposition and composition studies, we test target and off-target intervention effects for three externally defined empathy mechanisms and compare their span with persona-induced shifts.

\noindent\textbf{Empathy and persona-conditioned bias.}
Prior work on emotional-support generation studies how models produce supportive replies~\citep{liu2021esconv,sabour2022cem}. \citet{sharma2020epitome} introduced EPITOME, the ER/IP/EX framework we adopt. \citet{maurya2024multidim} argue for dimension-aware empathy evaluation, and \citet{welivita2024perceived} note that LLM responses can be perceived as more empathic than human ones, motivating our cross-evaluator comparison with an LLM judge~\citep{gao2025reliable,lee2025heart}. More broadly, persona prompting can steer model behavior and expose social biases~\citep{liu2024personasteered,li2024datapersonas}. \citet{kim2025empathygaps} document empathy gaps under perceiver/experiencer conditioning, and \citet{shen2025multidemo} propose dimension-specific metrics for multi-demographic personas. This line establishes that empathy varies with prompting and persona through response-level scoring, but cannot tell whether mechanisms share one direction or whether personas scale empathy uniformly. Our work addresses both by extracting mechanism vectors, testing their orthogonality, and projecting persona-conditioned activations onto them.

\begin{figure*}[t!]
\centering
\includegraphics[width=0.82\textwidth]{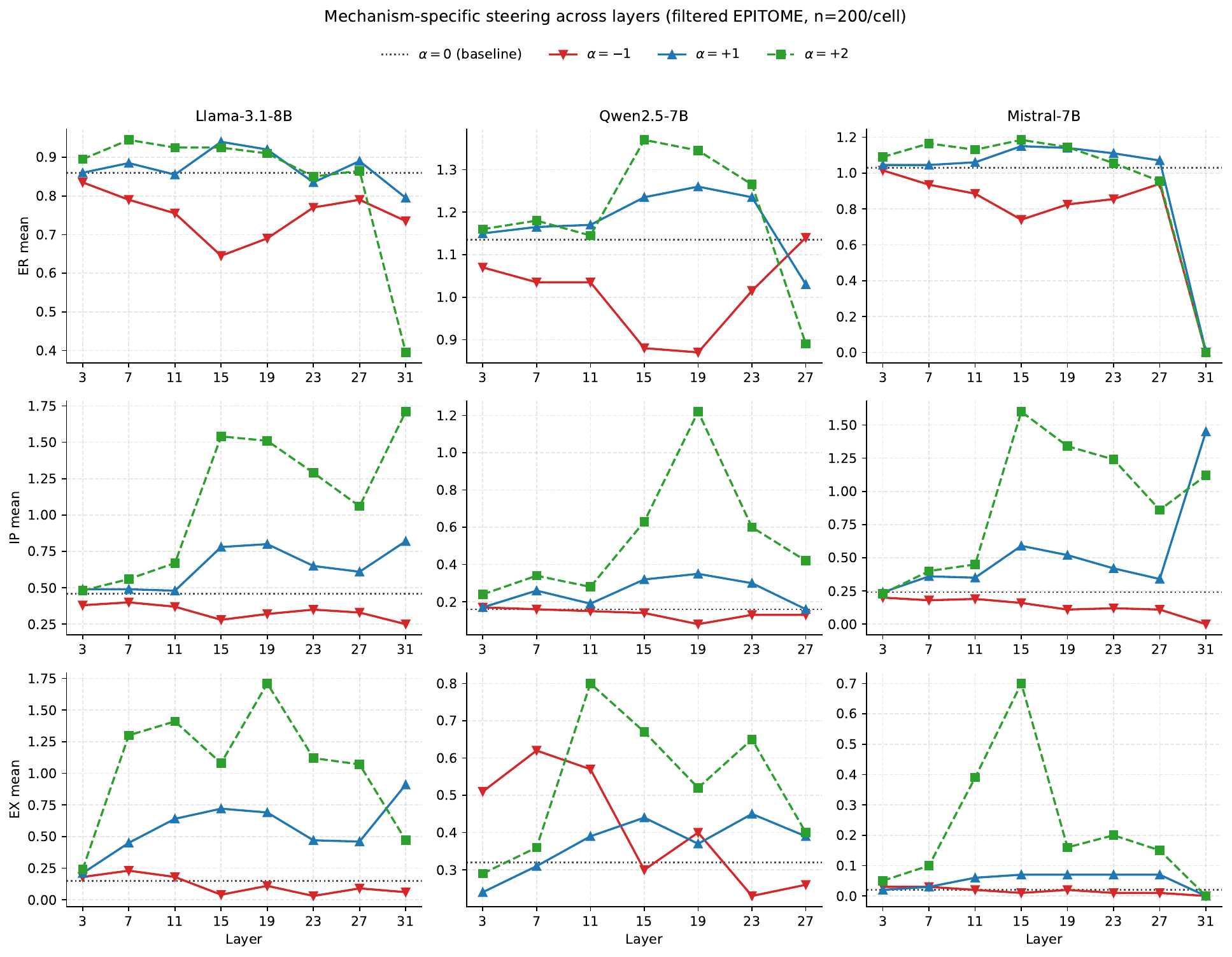}
\caption{%
Mean ER, IP, and EX levels under steering across layers (filtered EPITOME; $n=200$ per cell). Rows are mechanisms, columns are models; each panel plots $\alpha \in \{-1, +1, +2\}$ with $\alpha=0$ as a dotted baseline. Qwen has no layer-31 block.%
}
\label{fig:mechanism-lines}
\end{figure*}

\section{Methods}
\label{sec:methods}



\paragraph{Models and data.} Figure~\ref{fig:main-overview} summarizes the vector-extraction and inference-time steering pipeline. We evaluate three widely used instruction-tuned open models: \textbf{Llama-3.1-8B-Instruct}~\citep{dubey2024llama3} (32 blocks), \textbf{Qwen2.5-7B-Instruct}~\citep{qwen25,qwen2} (28 blocks), and \textbf{Mistral-7B-Instruct}~\citep{jiang2023mistral7b} (32 blocks). All steering vectors are built from filtered EPITOME data, with separate subsets for ER, IP, and EX. Before vector construction and evaluation, we remove seeker posts shorter than 100 characters and posts containing suicide or self-harm keywords: short posts often lack situational context for mechanism-level analysis, and self-harm posts are more likely to trigger refusal or safety-template behavior. The full keyword list and filtering script are released with the code. The EPITOME classifiers assign each generated response a mechanism score in $\{0,1,2\}$, which we treat as a learned proxy for the corresponding human annotation rather than as a direct measure of perceived empathy. The LLM-based judge uses the same ER/IP/EX annotation instructions as the original EPITOME work~\citep{sharma2020epitome}.


\paragraph{Mechanism-vector extraction.} For each mechanism $m \in \{\mathrm{ER}, \mathrm{IP}, \mathrm{EX}\}$, model, and layer, we build a Contrastive Activation Addition vector by contrasting the mean hidden states of a high-empathy pool $\mathcal{P}_m$ and a low-empathy pool $\mathcal{N}_m$ drawn from the EPITOME data for mechanism $m$. We use $|\mathcal{P}_m|=|\mathcal{N}_m|=100$ rows, where $\mathcal{P}_m$ contains rows with empathy label $y_m \ge 1$ and $\mathcal{N}_m$ contains rows with $y_m=0$. For each row $x$ (a seeker--response sequence), we run a teacher-forced forward pass and write $h_\ell(x)$ for the layer-$\ell$ residual-stream hidden state at the final response token. The mechanism vector is the difference of pool means:
\begin{equation}
v_\ell^m \;=\;
\frac{1}{|\mathcal{P}_m|}\!\sum_{x \in \mathcal{P}_m}\! h_\ell(x)
\;-\;
\frac{1}{|\mathcal{N}_m|}\!\sum_{x \in \mathcal{N}_m}\! h_\ell(x).
\label{eq:caa}
\end{equation}
The two pools are not paired by seeker post --- we contrast pool-level means rather than per-row differences.

\begin{table*}[t]
\centering
\small
\setlength{\tabcolsep}{3.5pt}
\begin{tabular}{ll|c|ccc|c|ccc}
\toprule
& & \multicolumn{4}{c|}{\textbf{EPITOME classifier}} & \multicolumn{4}{c}{\textbf{GPT-5-mini judge}} \\
& & \multicolumn{1}{c|}{\textbf{Base}} & \multicolumn{3}{c|}{\textbf{Steering change vs.\ base}} & \multicolumn{1}{c|}{\textbf{Base}} & \multicolumn{3}{c}{\textbf{Steering change vs.\ base}} \\
\textbf{Mechanism} & \textbf{Model} & $\alpha{=}0$ & $\Delta(-1)$ & $\Delta(+1)$ & $\Delta(+2)$ & $\alpha{=}0$ & $\Delta(-1)$ & $\Delta(+1)$ & $\Delta(+2)$ \\
\midrule
ER & Llama-3.1   & 0.860 & \tabneg{20}{0.215} & \tabpos{10}{0.080} & \tabpos{9}{0.065} & 1.670 & \tabneg{21}{0.220} & \tabpos{14}{0.135} & \tabpos{17}{0.175} \\
ER & Qwen2.5     & 1.135 & \tabneg{24}{0.255} & \tabpos{12}{0.100} & \tabpos{22}{0.235} & 1.690 & \tabneg{21}{0.225} & \tabpos{14}{0.135} & \tabpos{19}{0.200} \\
ER & Mistral-7B  & 1.030 & \tabneg{27}{0.290} & \tabpos{13}{0.120} & \tabpos{16}{0.155} & 1.600 & \tabneg{25}{0.270} & \tabpos{21}{0.225} & \tabpos{29}{0.315} \\
\midrule
EX & Llama-3.1   & 0.150 & \tabneg{12}{0.110} & \tabpos{47}{0.570} & \tabpos{70}{0.930} & 1.150 & \tabneg{15}{0.145} & \tabpos{25}{0.270} & \tabpos{30}{0.325} \\
EX & Qwen2.5     & 0.320 & \tabneg{6}{0.020} & \tabpos{13}{0.120} & \tabpos{32}{0.350} & 1.030 & \tabneg{11}{0.090} & \tabpos{12}{0.105} & \tabpos{22}{0.235} \\
EX & Mistral-7B  & 0.020 & \tabneg{5}{0.010} & \tabpos{8}{0.050} & \tabpos{55}{0.680} & 0.900 & \tabpos{6}{0.020} & \tabpos{10}{0.075} & \tabpos{18}{0.185} \\
\midrule
IP & Llama-3.1   & 0.460 & \tabneg{18}{0.180} & \tabpos{29}{0.320} & \tabpos{80}{1.080} & 1.875 & \tabneg{11}{0.085} & \tabpos{7}{0.030} & \tabpos{9}{0.065} \\
IP & Qwen2.5     & 0.160 & \tabneg{6}{0.020} & \tabpos{16}{0.160} & \tabpos{40}{0.470} & 1.775 & \tabpos{5}{0.005} & \tabneg{9}{0.060} & \tabneg{5}{0.005} \\
IP & Mistral-7B  & 0.240 & \tabneg{10}{0.080} & \tabpos{32}{0.350} & \tabpos{90}{1.360} & 1.750 & \tabpos{9}{0.065} & \tabpos{11}{0.085} & \tabpos{15}{0.150} \\
\bottomrule
\end{tabular}
\caption{Layer-15 mean ER/EX/IP levels under mechanism-targeted steering at $\alpha \in \{-1, 0, +1, +2\}$, on three models ($n=200$ per cell), scored by the EPITOME classifier and by a GPT-5-mini judge on the same generations. The $\alpha=0$ column is the unsteered baseline; \textbf{\textcolor{tablepostext}{green}} marks increases and \textbf{\textcolor{tablenegtext}{red}} marks decreases relative to it. Full per-layer EPITOME values are in  Table~\ref{tab:full-layer-sweep}.}
\label{tab:layer15-main}
\end{table*}

\paragraph{Layer sweep and evaluation.} We sweep layers $\ell = 3, 7, \ldots, 31$ in steps of 4 (omitting $\ell{=}31$ for Qwen, which has only 28 blocks) and test $\alpha \in \{-1,0,+1,+2\}$. At each generation step, a forward hook adds the selected mechanism vector to the current last-token hidden state:
\begin{equation}
h_{\ell,t} \leftarrow h_{\ell,t} + \alpha v_\ell^m .
\label{eq:steering}
\end{equation}

We generate one response per evaluation seeker and score the outputs using the EPITOME classifiers.
The layer-sweep and out-of-domain steering runs use sampled decoding with temperature
$T{=}0.8$, top-$p{=}0.9$. The exact generation prompt is listed in Appendix~\ref{sec:appendix-prompts}. For analyses requiring a single intervention layer, we use the canonical layer identified by the sweep; Section~\ref{sec:results-layer} shows that this corresponds to $\ell=15$. For this canonical-layer setting, we cross-validate the steering effects using two automated evaluators: the EPITOME classifiers and a GPT-5-mini judge using the same ER/IP/EX rubrics (Section~\ref{sec:results-aux}). We restrict the LLM judge to this canonical-layer subset --- all three mechanisms, all three models, and $\alpha \in \{-1,0,+1,+2\}$ --- because it is the main operating point used for cross-mechanism and persona analyses. Agreement between the two evaluators tests whether the steering directions are specific to the EPITOME classifiers. Prior work has shown that LLM-based evaluators correlate strongly with human raters on empathy-related tasks~\citep{gao2025reliable,lee2025heart}, supporting this cross-evaluator comparison. Given the $n=200$ evaluation size per cell, we report paired bootstrap confidence intervals and multi-seed checks for the canonical-layer setting in Section~\ref{sec:results-layer}. Our claims concern activation-level structure under fixed empathy proxies, not deployment-ready gains in perceived empathic support; extending them to perceived empathy quality would require human evaluation.


\paragraph{Persona and projection analysis.} For persona-conditioned generation we prepend \texttt{You are a \{persona\}.}, comparing each persona against the neutral \textit{person} baseline across control traits and social attributes. Throughout this analysis, $v_{\text{ER}}$, $v_{\text{IP}}$, and $v_{\text{EX}}$ denote the mechanism vectors at the canonical layer for the model being analyzed. For each persona and seeker, we generate a response, run a teacher-forced pass over the persona prompt and response, and mean-pool the generated-response token activations at the same layer into $a \in \mathbb{R}^d$. We measure alignment with the ER direction by the cosine
\begin{equation}
p_{\text{ER}} = \cos(a, v_{\text{ER}})
= \frac{\langle a, v_{\text{ER}} \rangle}{\|a\|\,\|v_{\text{ER}}\|}.
\label{eq:projection}
\end{equation}
We average this projection over the same 200 seekers for each persona and report the paired difference from the \textit{person} baseline as $\Delta p_{\text{ER}}$; the same procedure yields $\Delta p_{\text{IP}}$ and $\Delta p_{\text{EX}}$.


\paragraph{Mechanism vector residualization.} To measure how much the mechanism vectors $\{v_{\text{ER}},v_{\text{EX}},v_{\text{IP}}\}$ at the canonical layer share representational structure, we residualize each vector onto the orthogonal complement of the span of the other two and recompute pairwise cosines and steering effects. Concretely, we compute the least-squares residual of one vector after regressing it on the other two, then apply matched-norm scaling before steering. Because this projection uses the joint span of the other two vectors, it is invariant to their ordering. Full formulas and norm-ratio analyses are in Appendix~\ref{sec:appendix-residualization}.

\begin{table*}[!t]
\centering
\footnotesize
\setlength{\tabcolsep}{3pt}
\begin{tabular}{llc|rrr|rrr|rrr}
\toprule
& & & \multicolumn{3}{c|}{\textbf{Llama}} & \multicolumn{3}{c|}{\textbf{Qwen}} & \multicolumn{3}{c}{\textbf{Mistral}} \\
\textbf{Vector} & \textbf{Steering} & $\boldsymbol{\alpha}$ & $\Delta$ER & $\Delta$EX & $\Delta$IP & $\Delta$ER & $\Delta$EX & $\Delta$IP & $\Delta$ER & $\Delta$EX & $\Delta$IP \\
\midrule
\multirow{2}{*}{ER} & negative & $-1$ & $-$0.215 & $+$0.040 & $+$0.240 & $-$0.255 & $-$0.070 & $-$0.020 & $-$0.290 & $+$0.020 & $+$0.110 \\
                    & positive & $+1$ & $+$0.080 & $+$0.010 & $-$0.150 & $+$0.100 & $+$0.115 & $\phantom{+}$0.000 & $+$0.120 & $-$0.010 & $-$0.080 \\
\midrule
\multirow{2}{*}{EX} & negative & $-1$ & $-$0.005 & $-$0.110 & $+$0.320 & $-$0.155 & $-$0.020 & $+$0.020 & $\phantom{+}$0.000 & $-$0.010 & $+$0.260 \\
                    & positive & $+1$ & $-$0.225 & $+$0.570 & $-$0.060 & $-$0.015 & $+$0.120 & $-$0.020 & $-$0.040 & $+$0.050 & $-$0.110 \\
\midrule
\multirow{2}{*}{IP} & negative & $-1$ & $-$0.090 & $+$0.100 & $-$0.180 & $-$0.015 & $+$0.050 & $-$0.020 & $-$0.015 & $-$0.040 & $-$0.080 \\
                    & positive & $+1$ & $-$0.055 & $+$0.020 & $+$0.320 & $-$0.105 & $+$0.090 & $+$0.160 & $+$0.025 & $-$0.010 & $+$0.350 \\
\bottomrule
\end{tabular}
\caption{Layer-15 cross-mechanism EPITOME deltas under mechanism-targeted steering at $\alpha \in \{-1,+1\}$. Diagonal cells are target-mechanism effects for the named vector.}
\label{tab:cross-mechanism-layer15}
\end{table*}

\begin{table}[!t]
\centering
\small
\begin{tabular}{llrr}
\toprule
model & pair & original & residualized \\
\midrule
\multirow{3}{*}{Llama} 
& ER vs.\ EX & $+0.08$ & $+0.06$ \\
& ER vs.\ IP & $-0.29$ & $+0.29$ \\
& EX vs.\ IP & $-0.45$ & $+0.45$ \\
\midrule
\multirow{3}{*}{Qwen} 
& ER vs.\ EX & $+0.13$ & $+0.03$ \\
& ER vs.\ IP & $-0.32$ & $+0.30$ \\
& EX vs.\ IP & $-0.46$ & $+0.45$ \\
\midrule
\multirow{3}{*}{Mistral} 
& ER vs.\ EX & $+0.09$ & $+0.05$ \\
& ER vs.\ IP & $-0.25$ & $+0.24$ \\
& EX vs.\ IP & $-0.53$ & $+0.53$ \\
\bottomrule
\end{tabular}
\caption{Layer-15 pairwise cosines between mechanism vectors before and after residualizing each vector against the other two.}
\label{tab:geometry}
\end{table}

\section{Results}
\label{sec:results}


Table~\ref{tab:layer15-main} summarizes the canonical layer-15 steering effects; we next explain how that layer is selected and how the mechanism directions interact.


\subsection{Localizing Mechanism Steering}
\label{sec:results-layer}

Figure~\ref{fig:mechanism-lines} plots mean ER/IP/EX levels across layers. The sweep identifies a stable middle-layer operating point: at $\alpha{=}{+1}$, layer 15 yields positive target-mechanism steering for every $(\text{model}, \text{mechanism})$ cell, whereas later layers are more model-dependent and the final layer is unstable. On Llama, all layer-15 sign-direction results survive multi-seed and decoding-seed checks (Appendix~\ref{sec:appendix-robustness}).

\paragraph{Robustness and Transfer Checks.}
\label{sec:results-aux}
Three additional checks support the layer-15 effect: the GPT-5-mini judge in Table~\ref{tab:layer15-main} preserves the ER/EX directional pattern, layer-15 ER vectors partially transfer to EmpatheticDialogues~\citep{rashkin2019empatheticdialogues} (Table~\ref{tab:ood-er}), and WikiText-103 perplexity rises modestly at $\alpha \in \{-1,+1\}$ but more at $\alpha{=}{+2}$ (Table~\ref{tab:wikitext-ppl}).

\subsection{Sub-Mechanism Trade-Offs and Geometry}
\label{sec:results-geometry}


\paragraph{Cross-mechanism spillovers.} 
Table~\ref{tab:cross-mechanism-layer15} summarizes the layer-15 target and cross-mechanism deltas. Each delta is measured against the matching $\alpha{=}0$ baseline from the same mechanism-vector run; diagonal cells are target effects and off-diagonal cells are non-target effects. All target effects move in the expected direction across the three models, but the non-target effects show that control is not selective. 
The clearest cross-model pattern appears under EX steering: $\alpha{=}{+}1$ raises EX while ER and IP decrease on all three models (Llama $\Delta\mathrm{EX}{=}{+}0.57$, $\Delta\mathrm{ER}{=}{-}0.225$, with smaller but same-signed effects on Qwen and Mistral), and $\alpha{=}{-}1$ lowers EX while IP rises, most clearly on Llama and Mistral. A related pattern appears under IP steering: $\alpha{=}{+}1$ raises IP by $0.16$--$0.35$ while ER moves in the opposite direction on Llama and Qwen, consistent with the ER--IP anti-alignment we observe geometrically. ER $\alpha{=}{+}1$ lowers IP on Llama and Mistral but not Qwen, so we treat the ER--IP trade-off as model-dependent.

\paragraph{Geometric overlap.} At layer~15, the three mechanism vectors do not form cleanly separable directions on any model (Table~\ref{tab:geometry}). ER and EX are nearly orthogonal across models (cosines $+0.08$ to $+0.13$), while ER--IP ($-0.25$ to $-0.32$) and EX--IP ($-0.45$ to $-0.53$) are anti-aligned with the same sign pattern on Llama, Qwen, and Mistral. These cosine signs predict the \textit{direction} of the IP-involving spillovers in Table~\ref{tab:cross-mechanism-layer15} but not their magnitudes, which additionally reflect classifier saturation and token-level competition. The one apparent inconsistency---ER and EX nearly orthogonal yet EX $\alpha{=}{+}1$ consistently lowers ER---may reflect token-level competition, where question-form EX tokens displace ER acknowledgment. On Llama at $\alpha=+1$, residualization preserves ER steering while retaining $84\%$ of the original EX effect and $50\%$ of the original IP effect, showing that shared components contribute to steering behavior (Appendix~\ref{sec:appendix-residualization}). The three mechanisms therefore share representational structure rather than forming independent axes.

\subsection{Prompt Baseline}
\label{sec:results-prompt-baseline}



Can the mechanism-specific control of Section~\ref{sec:results-geometry} be approximated by prompting alone? We compare four prompt conditions on the same 200 seekers used for the ER steering run, with no activation intervention (Appendix~\ref{sec:appendix-prompt-baseline}). None cleanly isolates a single mechanism: P1/P2 shift multiple dimensions, while P3 strongly boosts EX but not IP and shortens Qwen/Mistral responses. In this set, prompting shifts the overall empathy profile rather than isolating one mechanism. 

\subsection{Persona Effects}
\label{sec:results-persona}


We measure how persona prompts (Table~\ref{tab:persona}) shift the EPITOME mean on Llama-3.1-8B-Instruct ($n=200$ per persona; full cross-model table in Appendix~\ref{sec:appendix-persona}). Personas move ER and IP labels substantially while leaving EX largely flat. Two personas show an ER--IP \emph{decoupling} on Llama:
\textit{black person} and \textit{engineer} both lower ER while raising IP. Because this decoupling does not replicate across models, we treat it as Llama-specific rather than a general property of persona prompting.


\begin{table}[t]
\centering
\small
\setlength{\tabcolsep}{4pt}
\begin{tabular}{lrrr}
\toprule
persona & $\Delta$ER & $\Delta$IP & $\Delta$EX \\
\midrule
\textit{empathetic person} & $+$0.175 & $-$0.25 & \textbf{$-$0.11} \\
\textit{cynical person}    & \textbf{$-$0.550} & $-$0.33 & $+$0.06 \\
\textit{white person}      & $-$0.040 & $-$0.05 & $+$0.09 \\
\textit{black person}      & $-$0.105 & \textbf{$+$0.48} & $\phantom{+}$0.00 \\
\textit{female}            & $+$0.030 & $+$0.07 & $-$0.05 \\
\textit{male}              & $-$0.070 & $+$0.02 & $+$0.03 \\
\textit{psychotherapist}   & $-$0.080 & $-$0.27 & $-$0.01 \\
\textit{engineer}          & $-$0.175 & $+$0.21 & $\phantom{+}$0.00 \\
\textit{Democratic supporter} & $+$0.020 & $-$0.24 & $-$0.01 \\
\textit{Republican supporter} & $-$0.035 & $-$0.35 & $+$0.03 \\
\bottomrule
\end{tabular}
\caption{Persona effects on Llama-3.1-8B (deltas vs.\ \textit{person}). Bold marks the largest absolute delta per column. Full per-model tables are in Appendix~\ref{sec:appendix-persona}.}
\label{tab:persona}
\end{table}

\subsection{Layer-15 Projection and a Race-Conditioned Steering Check}
\label{sec:results-projection}

\begin{figure}[t!]
\centering
\includegraphics[width=0.5\textwidth]{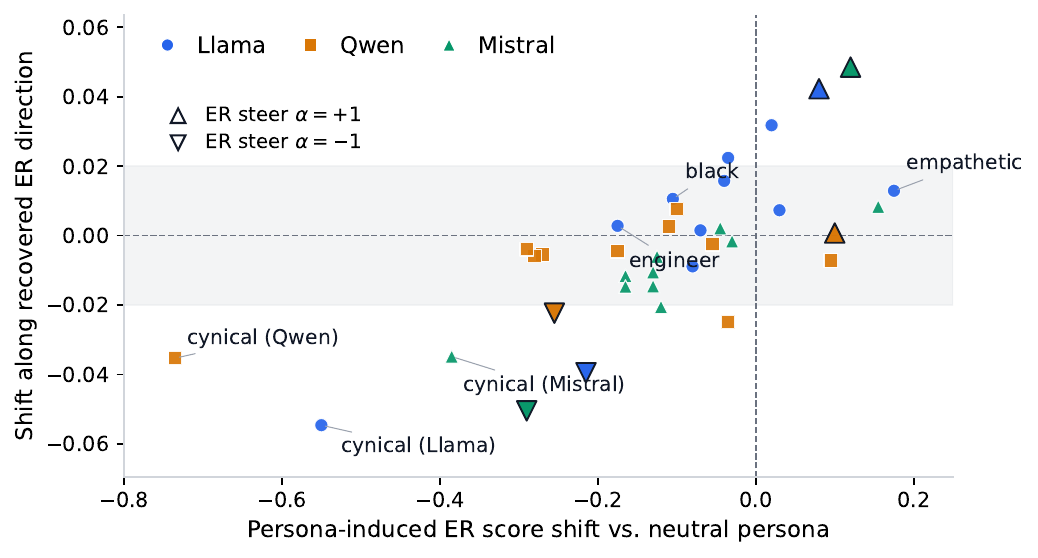}
\caption{Persona-induced ER shifts align weakly with the recovered ER direction. Small colored markers show persona prompts for each model; large outlined triangles show direct ER-steering controls.}
\label{fig:persona-projection}
\end{figure}

We ask a sharper question: do these persona-induced ER score changes correspond to movement along the recovered ER direction? We focus this projection analysis on $v_{\text{ER}}$ because ER is the most direct EPITOME mechanism for emotional acknowledgment and gives the clearest interpretation for the \textit{empathetic}/\textit{cynical} control personas. This choice does not imply that ER is always the largest persona effect. IP projections are reported in Appendix~\ref{sec:appendix-projection-full}, and the subspace analysis below tests the full ER/IP/EX mechanism basis. Across two control personas (\textit{empathetic}, \textit{cynical}) and eight social personas, we project layer-15 activations of persona-conditioned generations onto $v_{\text{ER}}$ and compare the resulting $\Delta p_{\text{ER}}$ to the EPITOME $\Delta$ER. Figure~\ref{fig:persona-projection} shows the diagnostic pattern: persona-induced ER changes often remain near $\Delta p_{\text{ER}}{=}0$, unlike direct $v_{\text{ER}}$ steering.


\paragraph{Projection analysis.} Two observations follow. First, label shifts are large but projection shifts are small: \textit{cynical} produces the largest ER decreases across models, reaching $\lvert\Delta\mathrm{ER}\rvert=0.74$, while the largest persona-induced projection shift is only $\lvert\Delta p_{\text{ER}}\rvert=0.055$. Second, direct ER-steering controls provide a reference for movement along $v_{\text{ER}}$: on Llama and Mistral, steering moves ER scores and ER projection together, whereas persona prompts can produce large ER-score shifts with weaker projection movement. Qwen's projection response is weaker even under direct steering, making this calibration less informative on Qwen. Overall, the recovered $v_{\text{ER}}$ captures only a small directional component of persona-associated ER shifts.


\paragraph{Subspace decomposition.} The mismatch is not specific to the ER axis. We measure how much of each layer-15 persona-mean activation shift (relative to the \textit{person} baseline) is captured by the orthonormalized span of $\{v_{\text{ER}},v_{\text{IP}},v_{\text{EX}}\}$ (formal definition in Appendix~\ref{sec:appendix-persona-subspace}). Averaged across personas, this fraction is only $2.6$--$3.1\%$ across models, leaving roughly $97\%$ of squared shift magnitude outside the recovered mechanism basis. The residual remains structured: its top five PCs explain $27$--$30\%$ of residual variance. Persona prompting therefore recruits structured representational variation largely outside the mechanism basis recovered by contrastive empathy steering.



\paragraph{Race-conditioned steering check.} As a behavioral counterpart to the projection result, we apply $v_{\text{ER}}$ to three race-conditioned personas. $\alpha{=}{+1}$ raises ER for all three, but does not equalize the \textit{black person}--\textit{person} ER gap; we report this as a single-model behavioral check rather than a general fairness claim (Appendix~\ref{sec:appendix-persona-steer}).

\section{Discussion}
\label{sec:discussion}

\subsection{Layer Localization and Mechanism Geometry}

Empathy in our setup is steerable in middle layers, consistent with prior CAA results~\citep{panickssery2024steering}. The last layer is unstable in two of three models even at moderate $\alpha$, possibly reflecting tighter coupling between final residual states and next-token logits.


The three EPITOME-anchored vectors share functional components: their geometric overlap matches behavioral spillovers, and residualizing away shared components weakens control (Appendix~\ref{sec:appendix-residualization}). The prompt baseline reinforces this: hand-crafted wording shifts the overall empathy profile rather than isolating a single mechanism. Together with our length/question-control analysis, this leads us to interpret EX steering as classifier-level control rather than evidence of deeper exploratory reasoning.


\subsection{Persona Effects: A Mechanistic Analysis}

Persona prompts move ER labels substantially, but those shifts are only weakly aligned with the recovered $v_{\text{ER}}$ (Section~\ref{sec:results-projection}). The stronger result is a subspace result: at the layer-15 response-token representation, the recovered ER/IP/EX basis captures only about $3\%$ of persona-mean squared activation-shift magnitude across models, while the residual shifts retain low-rank structure. This suggests that mechanism-targeted steering and persona prompting occupy different representational regimes: CAA vectors give reliable local control over EPITOME mechanisms, but persona prompts recruit broader structured variation that the recovered mechanism basis does not span. The contrast between \textit{cynical} --- where projection and label move together most clearly --- and most other personas suggests that explicit trait-level ER suppression is partly recoverable, while demographic and occupational shifts are not captured by the recovered ER direction.



\paragraph{Structure in the residual subspace.}
The residual variation is not diffuse noise: at the layer-15 response-token representation, after removing the ER/IP/EX mechanism-subspace component, the top two residual PCs explain $17$--$20\%$ of residual variance and the top five explain $27$--$30\%$. This suggests that persona prompting recruits structured representational components beyond the EPITOME-anchored basis. What those components encode --- persona-specific style, social-category associations, or interactions with other attributes --- remains an open question. The structure itself, however, suggests that persona effects admit a principled decomposition, motivating future persona-aware controllers that target residual structure rather than only the recovered mechanism directions.

\section{Conclusion}
\label{sec:conclusion}



We examined whether candidate directions derived from EPITOME's ER/IP/EX labels produce distinguishable intervention effects and how much structure they share, and whether persona prompts shift empathy along the recovered directions. 
Layer 15 provides a reliable cross-model intervention point, while hand-crafted prompts do not isolate any mechanism. Persona-induced empathy shifts are weakly aligned with the recovered directions, whose span captures little of the activation changes. 
These findings distinguish steerability from selective control: target scores shift consistently across models, but steering one dimension changes the others. Under this EPITOME-based definition, expressed empathy is steerable but multi-axial, and the recovered directions do not form a complete control basis for persona-conditioned changes.  More broadly, dimensions used to evaluate supportive empathy need not provide independent intervention targets.
Future work should explore multi-vector controllers and persona-conditioned directions with  robustness checks across models and support domains, and human evaluation is needed to verify whether proxy-level ER/IP/EX shifts correspond to perceived empathic support.  


\section{Limitations}
\label{sec:limitations}

\paragraph{Proxy evaluation.}
Our evaluation remains proxy-based: EPITOME and GPT-5-mini scores are learned proxies for human empathy annotations rather than direct measurements of perceived empathy, and we do not collect human annotations of steered or persona-conditioned generations. Our central claims concern representation-level structure under fixed automated proxies; human evaluation is required before interpreting these shifts as improvements in perceived empathic support. The EX/IP classifiers also behave effectively as 0/2 binaries, possibly inflating the apparent magnitude of steering. We report only a lightweight length/question-control analysis, so response-style confounds remain possible, especially for EX, which should be read as classifier-level control rather than evidence of deeper exploratory reasoning.

\paragraph{Robustness scope.}
Our seed checks are limited to the canonical Llama layer-15 setting: we vary vector-extraction/sample seeds and separately test decoding seeds with the vector/evaluation split fixed, but do not extend either check to Qwen or Mistral. Persona-conditioned projection results are single-run estimates over 200 seekers per persona. The persona set is narrow and US-centric; these prompts are controlled probes rather than representative demographic samples, so we do not make population-level fairness claims, and non-US and intersectional personas remain future work. The prompt baseline uses four hand-crafted instructions and serves only as a scale reference, not a direct head-to-head superiority test; systematically optimized prompting remains future work.

\paragraph{Method scope.}
The residual-PC analysis of persona shifts is descriptive: it shows low-rank structure after removing the recovered ER/IP/EX mechanism subspace, but does not identify a causal persona mechanism. Our steering vectors are static layer-15 CAA directions on the residual stream; adaptive, multi-vector, or sub-module-level controllers, as well as persona-conditioned vectors, are left to future work. We also hypothesize but do not verify (e.g., via logit-lens) that final-layer instability reflects tighter coupling to next-token logits.

\section{Ethical Considerations}
\label{sec:ethics}

Persona categories are used only for representational analysis; reported empathy differences are properties of the model under our prompts, not of the referenced groups. The race-conditioned steering check is not a fairness intervention, and we do not recommend single-vector steering for debiasing. We use previously released EPITOME Reddit support data~\citep{sharma2020epitome} and filter short or self-harm-related posts to reduce high-risk support contexts. Because the same directions can lower empathic phrasing at $\alpha{<}0$, real deployment would require human evaluation, safety review, and safeguards.

\section*{Acknowledgments}

This research was partly supported by the Institute for Information \& Communications Technology Planning \& Evaluation (IITP) under the Artificial Intelligence Convergence Innovation Human Resources Development grant [IITP-2026-RS-2023-00255968] and the ITRC (Information Technology Research Center) support program [IITP-2026-RS-2021-II212051]. This work was also supported by the National Research Foundation of Korea (NRF) grant funded by the Korea government (MSIT) (RS-2026-25469593).

\bibliography{custom}

\appendix

\section{Full Layer Sweep Table}
\label{sec:appendix-fullsweep}

Table~\ref{tab:full-layer-sweep} reports the per-cell mean scores summarized by Figure~\ref{fig:mechanism-lines} in the main text. The figure shows the $\alpha \in \{-1, +1, +2\}$ slices over a constant $\alpha=0$ baseline; this table gives all four $\alpha$ values per layer for ER, EX, and IP.

\begin{table*}[h]
\centering
\scriptsize
\setlength{\tabcolsep}{2.5pt}
\resizebox{\textwidth}{!}{
\begin{tabular}{ll|cccc|cccc|cccc}
\toprule
& & \multicolumn{4}{c|}{\textbf{Llama}} & \multicolumn{4}{c|}{\textbf{Qwen}} & \multicolumn{4}{c}{\textbf{Mistral}} \\
\textbf{Metric} & $\ell$ & $-$1 & 0 & $+$1 & $+$2 & $-$1 & 0 & $+$1 & $+$2 & $-$1 & 0 & $+$1 & $+$2 \\
\midrule
\multirow{8}{*}{ER} & 3 & 0.83 & 0.86 & 0.86 & 0.90 & 1.07 & 1.14 & 1.15 & 1.16 & 1.01 & 1.03 & 1.04 & 1.09 \\
 & 7 & 0.79 & 0.86 & 0.89 & 0.94 & 1.03 & 1.14 & 1.17 & 1.18 & 0.94 & 1.03 & 1.04 & 1.17 \\
 & 11 & 0.76 & 0.86 & 0.85 & 0.93 & 1.03 & 1.14 & 1.17 & 1.15 & 0.89 & 1.03 & 1.06 & 1.13 \\
 & 15 & 0.65 & 0.86 & 0.94 & 0.93 & 0.88 & 1.14 & 1.24 & 1.37 & 0.74 & 1.03 & 1.15 & 1.19 \\
 & 19 & 0.69 & 0.86 & 0.92 & 0.91 & 0.87 & 1.14 & 1.26 & 1.34 & 0.82 & 1.03 & 1.14 & 1.15 \\
 & 23 & 0.77 & 0.86 & 0.83 & 0.85 & 1.01 & 1.14 & 1.24 & 1.26 & 0.85 & 1.03 & 1.11 & 1.05 \\
 & 27 & 0.79 & 0.86 & 0.89 & 0.86 & 1.14 & 1.14 & 1.03 & 0.89 & 0.94 & 1.03 & 1.07 & 0.95 \\
 & 31 & 0.73 & 0.86 & 0.80 & 0.40 & --- & --- & --- & --- & 0.00 & 1.03 & 0.01 & 0.00 \\
\midrule
\multirow{8}{*}{EX} & 3 & 0.18 & 0.15 & 0.21 & 0.24 & 0.51 & 0.32 & 0.24 & 0.29 & 0.03 & 0.02 & 0.02 & 0.05 \\
 & 7 & 0.23 & 0.15 & 0.45 & 1.30 & 0.62 & 0.32 & 0.31 & 0.36 & 0.03 & 0.02 & 0.03 & 0.10 \\
 & 11 & 0.18 & 0.15 & 0.64 & 1.41 & 0.57 & 0.32 & 0.39 & 0.80 & 0.02 & 0.02 & 0.06 & 0.39 \\
 & 15 & 0.04 & 0.15 & 0.72 & 1.08 & 0.30 & 0.32 & 0.44 & 0.67 & 0.01 & 0.02 & 0.07 & 0.70 \\
 & 19 & 0.11 & 0.15 & 0.69 & 1.71 & 0.40 & 0.32 & 0.37 & 0.52 & 0.02 & 0.02 & 0.07 & 0.16 \\
 & 23 & 0.03 & 0.15 & 0.47 & 1.12 & 0.23 & 0.32 & 0.45 & 0.65 & 0.01 & 0.02 & 0.07 & 0.20 \\
 & 27 & 0.09 & 0.15 & 0.46 & 1.07 & 0.26 & 0.32 & 0.39 & 0.40 & 0.01 & 0.02 & 0.07 & 0.15 \\
 & 31 & 0.06 & 0.15 & 0.91 & 0.47 & --- & --- & --- & --- & 0.00 & 0.02 & 0.00 & 0.00 \\
\midrule
\multirow{8}{*}{IP} & 3 & 0.38 & 0.46 & 0.49 & 0.48 & 0.17 & 0.16 & 0.17 & 0.24 & 0.20 & 0.24 & 0.24 & 0.23 \\
 & 7 & 0.40 & 0.46 & 0.49 & 0.56 & 0.16 & 0.16 & 0.26 & 0.34 & 0.18 & 0.24 & 0.36 & 0.40 \\
 & 11 & 0.37 & 0.46 & 0.48 & 0.67 & 0.15 & 0.16 & 0.19 & 0.28 & 0.19 & 0.24 & 0.35 & 0.45 \\
 & 15 & 0.28 & 0.46 & 0.78 & 1.54 & 0.14 & 0.16 & 0.32 & 0.63 & 0.16 & 0.24 & 0.59 & 1.60 \\
 & 19 & 0.32 & 0.46 & 0.80 & 1.51 & 0.08 & 0.16 & 0.35 & 1.22 & 0.11 & 0.24 & 0.52 & 1.34 \\
 & 23 & 0.35 & 0.46 & 0.65 & 1.29 & 0.13 & 0.16 & 0.30 & 0.60 & 0.12 & 0.24 & 0.42 & 1.24 \\
 & 27 & 0.33 & 0.46 & 0.61 & 1.06 & 0.13 & 0.16 & 0.16 & 0.42 & 0.11 & 0.24 & 0.34 & 0.86 \\
 & 31 & 0.25 & 0.46 & 0.82 & 1.71 & --- & --- & --- & --- & 0.00 & 0.24 & 1.45 & 1.12 \\
\bottomrule
\end{tabular}
}
\caption{Full layer sweep results: ER\_mean, EX\_mean, and IP\_mean across $\ell \in \{3,7,11,15,19,23,27,31\}$ and $\alpha \in \{-1,0,+1,+2\}$ ($n=200$ per cell). Qwen has no layer-31 block, shown as --- .}
\label{tab:full-layer-sweep}
\end{table*}

\section{Out-of-Domain Transfer (Full Table)}
\label{sec:appendix-ood}

We test domain transfer of the layer-15 ER vector by reusing each model's vector without re-estimation and replacing the evaluation inputs with situations from the EmpatheticDialogues (ED) test split~\citep{rashkin2019empatheticdialogues}. We restrict to seven negative-emotion categories (\textit{afraid}, \textit{angry}, \textit{anxious}, \textit{sad}, \textit{devastated}, \textit{lonely}, \textit{terrified}) to match the distress-focused context of the EPITOME source domain, deduplicate by conversation, and sample 200 situations from the \texttt{prompt} field; the ED prompt template appears in Appendix~\ref{sec:appendix-prompts}.

Table~\ref{tab:ood-er} reports the per-$\alpha$ ER means and $\Delta_{\alpha 0}$. The steering direction is preserved in every $(\text{model}, \alpha)$ cell, but the positive-direction magnitude varies substantially across models, while the negative direction transfers more uniformly. We therefore treat this result as limited directional-transfer evidence rather than evidence of domain-general ER control.

\begin{table}[h]
\centering
\scriptsize
\setlength{\tabcolsep}{4pt}
\resizebox{\columnwidth}{!}{
\begin{tabular}{l|rrrr|rrr}
\toprule
& \multicolumn{4}{c|}{\textbf{ER mean}} & \multicolumn{3}{c}{\textbf{$\Delta_{\alpha 0}$}} \\
\textbf{Model} & $\alpha{=}{-1}$ & $\alpha{=}0$ & $\alpha{=}{+1}$ & $\alpha{=}{+2}$ & $\alpha{=}{-1}$ & $\alpha{=}{+1}$ & $\alpha{=}{+2}$ \\
\midrule
Llama-3.1  & 0.665 & 0.915 & 0.935 & 1.030 & $-$0.250 & $+$0.020 & $+$0.115 \\
Qwen2.5    & 0.750 & 1.040 & 1.335 & 1.405 & $-$0.290 & $+$0.295 & $+$0.365 \\
Mistral-7B & 0.495 & 0.790 & 1.050 & 1.185 & $-$0.295 & $+$0.260 & $+$0.395 \\
\bottomrule
\end{tabular}
}
\caption{Out-of-domain ER transfer on EmpatheticDialogues (layer 15, $n=200$ per model). Directionality is preserved; the negative direction transfers uniformly across models, while the positive direction varies.}
\label{tab:ood-er}
\end{table}

\section{Language Fluency under Steering (Full Table)}
\label{sec:appendix-ppl}

Table~\ref{tab:wikitext-ppl} reports the per-cell WikiText-103 perplexity under layer-15 empathy steering on Llama-3.1-8B-Instruct, summarized in the auxiliary robustness paragraph in Section~\ref{sec:results-aux}.

\begin{table}[h]
\centering
\small
\setlength{\tabcolsep}{5pt}
\begin{tabular}{lcrr}
\toprule
\textbf{Mechanism} & $\boldsymbol{\alpha}$ & \textbf{PPL} $\downarrow$ & $\boldsymbol{\Delta}$\textbf{PPL} \\
\midrule
baseline & --- & 8.04 & --- \\
\midrule
\multirow{3}{*}{ER} & $-1$ & 8.18 & $+0.15$ \\
                    & $+1$ & 8.29 & $+0.26$ \\
                    & $+2$ & 9.37 & $+1.33$ \\
\midrule
\multirow{3}{*}{EX} & $-1$ & 8.24 & $+0.20$ \\
                    & $+1$ & 8.40 & $+0.36$ \\
                    & $+2$ & 10.13 & $+2.09$ \\
\midrule
\multirow{3}{*}{IP} & $-1$ & 8.18 & $+0.14$ \\
                    & $+1$ & 8.14 & $+0.10$ \\
                    & $+2$ & 8.58 & $+0.54$ \\
\bottomrule
\end{tabular}
\caption{WikiText-103 perplexity under layer-15 empathy steering on Llama-3.1-8B-Instruct. $\Delta$PPL relative to unsteered baseline ($8.04$).}
\label{tab:wikitext-ppl}
\end{table}



\section{Residualization Details}
\label{sec:appendix-residualization}

This appendix specifies the residualization procedure used in Section~\ref{sec:results-geometry} and Table~\ref{tab:geometry}, and provides an analysis for the sign flip observed in the ER--IP and EX--IP pairs.

\paragraph{Procedure.} Let $v_a, v_b, v_c \in \mathbb{R}^d$ be the layer-15 mechanism vectors for one model (e.g., $v_a = v_{\text{ER}}, v_b = v_{\text{EX}}, v_c = v_{\text{IP}}$). For each vector $v_a$, we compute its residual after projecting onto the orthogonal complement of $\mathrm{span}(v_b, v_c)$:
\begin{equation}
v_a^{\perp} \;=\; v_a \;-\; V_{bc}V_{bc}^{+}v_a,
\label{eq:residualize}
\end{equation}
where $V_{bc} = [\,v_b \;\; v_c\,] \in \mathbb{R}^{d \times 2}$ and $V_{bc}^{+}$ denotes the Moore--Penrose pseudoinverse. This is the standard least-squares projection of $v_a$ onto $\mathrm{span}(v_b, v_c)^{\perp}$; it depends only on the subspace spanned by $\{v_b, v_c\}$, not on which of them is processed first, so it is order-invariant. Equivalently, $v_a^{\perp}$ is the residual of an OLS regression of $v_a$ on $\{v_b, v_c\}$. We apply Eq.~\ref{eq:residualize} symmetrically to $v_b$ (against $\{v_a, v_c\}$) and $v_c$ (against $\{v_a, v_b\}$), then rescale each residual to match the original vector norm before steering (the matched-norm setting in the code). We report pairwise cosines $\cos(v_a^{\perp}, v_b^{\perp})$ etc.\ in Table~\ref{tab:geometry}.

\begin{figure*}[t]
\centering
\includegraphics[width=0.85\textwidth]{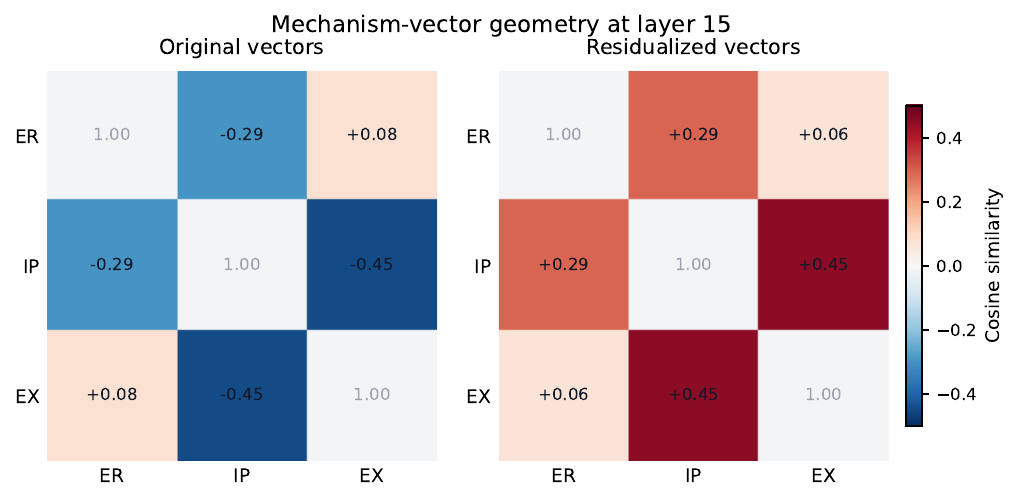}
\caption{Pairwise cosine between mechanism vectors at Llama layer 15, before (left) and after (right) residualizing each vector onto the orthogonal complement of the other two. The ER--IP and EX--IP sign flips reflect that the residualization removes different shared components from each vector; the residuals remain non-negligible in magnitude (see norm ratios above), so the residualized cosines should be read as an analysis of remaining alignment after shared-component removal rather than as evidence that residualization fails to disentangle the mechanisms. Values are computed from the Llama layer-15 vectors used in Section~\ref{sec:results-geometry}.}
\label{fig:cosine-heatmaps}
\end{figure*}

\paragraph{Sign-flip analysis.} The ER--IP and EX--IP cosines change sign after residualization ($-0.28 \to +0.25$ and $-0.43 \to +0.42$). This is expected when the three vectors share a substantial component: the projection removes different shared directions from each vector, so the residual pair can rotate even when the residualization procedure is order-invariant. At layer 15, the raw residual norms are still large relative to the originals ($\|v_{\text{ER}}^{\perp}\|/\|v_{\text{ER}}\| \approx 0.95$, $\|v_{\text{EX}}^{\perp}\|/\|v_{\text{EX}}\| \approx 0.89$, $\|v_{\text{IP}}^{\perp}\|/\|v_{\text{IP}}\| \approx 0.86$ before norm matching), so the sign flip is not driven by vanishing residual magnitude. We report both the cosine table and the residual-to-original norm ratio $r_a$ to make this explicit. Figure~\ref{fig:cosine-heatmaps} visualizes the pairwise geometry before and after residualization.

\paragraph{Steering with residualized vectors.} Re-running the layer-15 steering sweep with $v_a^{\perp}$ in place of $v_a$ preserves the sign of $\Delta_{\alpha 0}$ in all six (mechanism $\times$ direction) cells but reduces the magnitude on EX and IP. This is the behavioral counterpart of the geometric finding: removing the shared component leaves a usable but weaker mechanism-specific signal, supporting the main-text claim that ER, EX, and IP do not form independent axes.

\paragraph{Retention analysis.} Table~\ref{tab:residual-retention} quantifies how much of the $\alpha=+1$ steering effect remains after residualization on Llama. Although the raw residuals retain substantial norm (0.85--0.95 of the original), behavioral retention differs sharply across mechanisms: ER is preserved, EX is moderately reduced, and IP drops to half of its original effect. The retention spread (50--131\%) is much wider than the norm-ratio spread (0.85--0.95), so the shared component contributes to behavior beyond what its geometric magnitude predicts. The ER retention exceeding $100\%$ should be read as ``preserved'' rather than ``enhanced'': the original ER $\Delta_{\alpha=+1} = +0.080$ is the smallest among the three mechanisms.

\begin{table}[h]
\centering
\scriptsize
\setlength{\tabcolsep}{2.5pt}
\begin{tabular}{lrrrr}
\toprule
Mech. & Orig.\ $\Delta$ & Resid.\ $\Delta$ & Ret. & $\|v^{\perp}\|/\|v\|$ \\
\midrule
ER & $+0.080$ & $+0.105$ & $131.3\%$ & $0.954$ \\
EX & $+0.570$ & $+0.480$ & $\phantom{0}84.2\%$ & $0.892$ \\
IP & $+0.320$ & $+0.160$ & $\phantom{0}50.0\%$ & $0.855$ \\
\bottomrule
\end{tabular}
\caption{Layer-15 residualized steering retention on Llama at $\alpha=+1$. Original and residualized deltas are measured against the corresponding $\alpha=0$ baseline; retention is the residualized $\Delta$ divided by the original $\Delta$. The norm ratio is the raw residual norm before matched-norm scaling.}
\label{tab:residual-retention}
\end{table}

\section{Non-Target Label-Composition Control}
\label{sec:appendix-label-control}

We test whether the observed target effects and cross-mechanism coupling can be explained by non-target label imbalance in the extraction pools. On Llama at layer 15, we compare exact-matched pools, in which positive and negative responses have identical joint 0/1/2 counts on both non-target labels, with unmatched pools drawn from the same cleaned candidate set. The two arms use the same target-label counts, 200 evaluation seekers, and decoding settings; their $\alpha=0$ generations reproduce exactly. Table~\ref{tab:label-composition-control} reports the target effects before and after exact matching.

Matching reduces ER suppression under EX steering from $-0.12$ to $-0.05$ (paired difference $+0.07$, 95\% CI [$0.005$, $0.135$]). The ER--IP and EX--IP cosine estimates also weaken from $-0.395$ to $-0.124$ and from $-0.406$ to $-0.137$, respectively; these are point estimates because extraction-pool bootstrap intervals were not computed. Thus, non-target label imbalance is not necessary for target steerability, although pool composition contributes to cross-dimension coupling. This one-seed, Llama-only sensitivity analysis does not establish empathy specificity or rule out other dataset or pipeline effects.

\begin{table*}[t]
\centering
\small
\setlength{\tabcolsep}{5pt}
\begin{tabular}{lccc}
\toprule
Vector & Unmatched target $\Delta$ [95\% CI] & Exact-matched target $\Delta$ [95\% CI] & Matched $-$ unmatched [95\% CI] \\
\midrule
ER & $+0.030$ [$-0.040$, $0.100$] & $+0.065$ [$0.005$, $0.125$] & $+0.035$ [$-0.030$, $0.100$] \\
EX & $+0.680$ [$0.540$, $0.830$] & $+0.610$ [$0.460$, $0.760$] & $-0.070$ [$-0.230$, $0.090$] \\
IP & $+0.290$ [$0.140$, $0.440$] & $+0.270$ [$0.130$, $0.410$] & $-0.020$ [$-0.170$, $0.130$] \\
\bottomrule
\end{tabular}
\caption{Same-support non-target-label control on Llama at layer 15 ($\alpha=+1$ vs. $\alpha=0$, $n=200$, 10,000 evaluation-seeker bootstrap samples). Exact matching preserves positive target effects, and none of the target-effect differences between arms is significant.}
\label{tab:label-composition-control}
\end{table*}

\section{Full Persona Effects Table}
\label{sec:appendix-persona}

Table~\ref{tab:persona-full-cross} reports $\Delta$ER, $\Delta$IP, and $\Delta$EX (vs.\ the \textit{person} baseline) for all 11 personas across Llama-3.1-8B-Instruct, Qwen2.5-7B-Instruct, and Mistral-7B-Instruct-v0.3. Values are taken from the per-model persona-effects \texttt{summary\_vs\_person.csv} files ($n=200$ per persona). The Llama column is the source for the condensed table in the main text (Table~\ref{tab:persona}).

\begin{table*}[h]
\centering
\small
\setlength{\tabcolsep}{3.5pt}
\begin{tabular}{ll|rrr|rrr|rrr}
\toprule
& & \multicolumn{3}{c|}{\textbf{Llama}} & \multicolumn{3}{c|}{\textbf{Qwen}} & \multicolumn{3}{c}{\textbf{Mistral}} \\
category & persona & $\Delta$ER & $\Delta$IP & $\Delta$EX & $\Delta$ER & $\Delta$IP & $\Delta$EX & $\Delta$ER & $\Delta$IP & $\Delta$EX \\
\midrule
baseline & person & --- & --- & --- & --- & --- & --- & --- & --- & --- \\
\midrule
\multirow{2}{*}{personality}
 & \textit{empathetic person} & $+$0.175 & $-$0.250 & $-$0.110 & $+$0.095 & $+$0.170 & $-$0.040 & $+$0.155 & $+$0.050 & $-$0.020 \\
 & \textit{cynical person}    & $-$0.550 & $-$0.330 & $+$0.060 & $-$0.735 & $+$0.060 & $-$0.010 & $-$0.385 & $+$0.090 & $-$0.020 \\
\midrule
\multirow{2}{*}{gender}
 & \textit{female} & $+$0.030 & $+$0.070 & $-$0.050 & $-$0.055 & $+$0.010 & $-$0.010 & $-$0.030 & $+$0.040 & $-$0.010 \\
 & \textit{male}   & $-$0.070 & $+$0.020 & $+$0.030 & $-$0.280 & $-$0.060 & $+$0.010 & $-$0.165 & $+$0.030 & $\phantom{+}$0.000 \\
\midrule
\multirow{2}{*}{race}
 & \textit{white person} & $-$0.040 & $-$0.050 & $+$0.090 & $-$0.110 & $-$0.040 & $\phantom{+}$0.000 & $-$0.130 & $+$0.070 & $\phantom{+}$0.000 \\
 & \textit{black person} & $-$0.105 & $+$0.480 & $\phantom{+}$0.000 & $-$0.100 & $-$0.060 & $-$0.010 & $-$0.130 & $+$0.050 & $-$0.020 \\
\midrule
\multirow{2}{*}{occupation}
 & \textit{psychotherapist} & $-$0.080 & $-$0.270 & $-$0.010 & $-$0.035 & $+$0.200 & $-$0.010 & $-$0.120 & $+$0.040 & $-$0.020 \\
 & \textit{engineer}        & $-$0.175 & $+$0.210 & $\phantom{+}$0.000 & $-$0.270 & $-$0.090 & $-$0.030 & $-$0.125 & $-$0.010 & $-$0.020 \\
\midrule
\multirow{2}{*}{political}
 & \textit{Democratic supporter} & $+$0.020 & $-$0.240 & $-$0.010 & $-$0.175 & $+$0.030 & $-$0.030 & $-$0.045 & $-$0.030 & $-$0.010 \\
 & \textit{Republican supporter} & $-$0.035 & $-$0.350 & $+$0.030 & $-$0.290 & $-$0.100 & $\phantom{+}$0.000 & $-$0.165 & $+$0.020 & $-$0.020 \\
\bottomrule
\end{tabular}
\caption{Full per-model persona effects on EPITOME ER/IP/EX mean levels (deltas vs.\ \textit{person} baseline; $n=200$ per persona). The racial ER--IP decoupling pattern --- \textit{black person} raising IP while lowering ER --- is Llama-specific ($\Delta$IP$=+0.48$); Qwen shows $-0.06$ and Mistral $+0.05$ on the same persona. The personality control personas (\textit{empathetic}/\textit{cynical}) drive the largest ER shifts on every model.}
\label{tab:persona-full-cross}
\end{table*}

\section{Statistical Robustness Details}
\label{sec:appendix-robustness}

This appendix supports the statistical-robustness paragraph in Section~\ref{sec:results-layer}. We report (i) per-seed deltas for the multi-seed reproducibility check, (ii) decoding-seed deltas with a fixed vector/evaluation split, (iii) per-seed paired bootstrap CIs, and (iv) a persona-pair bootstrap comparison.

\paragraph{Per-seed deltas (multi-seed reproducibility).} Table~\ref{tab:seed-stability} reports $\Delta_{\alpha 0}$ for each (mechanism, $\alpha$) cell at Llama layer 15 across three independent vector-extraction/sample seeds. All six cells preserve the sign of the steering effect across all three seeds, with seed-to-seed standard deviations at most $0.082$.

\begin{table}[h]
\centering
\small
\setlength{\tabcolsep}{3pt}
\begin{tabular}{cc|rrr|r}
\toprule
mech. & $\alpha$ & $s{=}42$ & $s{=}52$ & $s{=}62$ & mean$\pm$sd \\
\midrule
\multirow{2}{*}{ER} & $-1$ & $-$0.215 & $-$0.210 & $-$0.130 & $-$0.185$\pm$0.05 \\
                    & $+1$ & $+$0.080 & $+$0.100 & $+$0.105 & $+$0.095$\pm$0.01 \\
\midrule
\multirow{2}{*}{EX} & $-1$ & $-$0.110 & $-$0.160 & $-$0.180 & $-$0.150$\pm$0.04 \\
                    & $+1$ & $+$0.570 & $+$0.560 & $+$0.660 & $+$0.597$\pm$0.06 \\
\midrule
\multirow{2}{*}{IP} & $-1$ & $-$0.180 & $-$0.150 & $-$0.190 & $-$0.173$\pm$0.02 \\
                    & $+1$ & $+$0.320 & $+$0.430 & $+$0.480 & $+$0.410$\pm$0.08 \\
\bottomrule
\end{tabular}
\caption{Seed stability of $\Delta_{\alpha 0}$ at Llama layer 15 across three independent vector-extraction/sample seeds ($n=200$ per cell). All 6 cells preserve sign across all 3 seeds.}
\label{tab:seed-stability}
\end{table}

\paragraph{Decoding-seed robustness.} We separately vary the decoding seed while keeping the layer-15 Llama vector/evaluation split fixed. Table~\ref{tab:decoding-seed-stability} reports the target-mechanism deltas for the canonical seed and two additional decoding seeds. All six mechanism--direction cells preserve sign across the three seeds; the standard deviation across decoding seeds ranges from $0.015$ to $0.081$.

\begin{table}[h]
\centering
\small
\setlength{\tabcolsep}{3pt}
\begin{tabular}{cc|rrr|r}
\toprule
mech. & $\alpha$ & $s{=}12$ & $s{=}22$ & $s{=}32$ & mean$\pm$sd \\
\midrule
\multirow{2}{*}{ER} & $-1$ & $-$0.215 & $-$0.200 & $-$0.250 & $-$0.222$\pm$0.026 \\
                    & $+1$ & $+$0.080 & $+$0.075 & $+$0.025 & $+$0.060$\pm$0.030 \\
\midrule
\multirow{2}{*}{EX} & $-1$ & $-$0.110 & $-$0.090 & $-$0.120 & $-$0.107$\pm$0.015 \\
                    & $+1$ & $+$0.570 & $+$0.450 & $+$0.420 & $+$0.480$\pm$0.079 \\
\midrule
\multirow{2}{*}{IP} & $-1$ & $-$0.180 & $-$0.120 & $-$0.280 & $-$0.193$\pm$0.081 \\
                    & $+1$ & $+$0.320 & $+$0.410 & $+$0.320 & $+$0.350$\pm$0.052 \\
\bottomrule
\end{tabular}
\caption{Decoding-seed stability of $\Delta_{\alpha 0}$ at Llama layer 15 with the vector/evaluation split fixed ($n=200$ per cell). All 6 cells preserve sign across decoding seeds $s \in \{12,22,32\}$.}
\label{tab:decoding-seed-stability}
\end{table}

\paragraph{Per-seed bootstrap CIs.}
Paired bootstrap intervals were computed by resampling the 200 evaluation seekers ($B=10{,}000$). All 18 mechanism--direction--seed intervals excluded zero and remained significant after Holm correction.

\paragraph{Persona-pair bootstrap.}
On Llama, the baseline ER difference between \textit{black person} and \textit{person} was $-0.105$ (95\% CI $[-0.180,-0.030]$), whereas the difference between \textit{white person} and \textit{person} was $-0.040$ (95\% CI $[-0.105,+0.020]$). We therefore do not infer similarity to the model's default from the non-significant \textit{white person} comparison.

\paragraph{Scope and caveats.} The bootstrap above quantifies sampling uncertainty over the $n=200$ evaluation set, Table~\ref{tab:seed-stability} quantifies procedural reproducibility across vector-extraction/sample seeds, and Table~\ref{tab:decoding-seed-stability} isolates decoding-seed variation with that split fixed. All robustness checks are conducted on Llama; extension to Qwen and Mistral is left to future work.

\section{Prompt Templates}
\label{sec:appendix-prompts}

Table~\ref{tab:prompt-templates} lists the prompt templates used throughout the paper, covering teacher-forced vector extraction, steering generation, OOD transfer on EmpatheticDialogues, and persona-conditioned generation. Extraction, steering, and OOD inputs are tokenized as raw text, whereas persona and prompt-baseline inputs use each model's chat template, which may add control tokens or default wrapper text. No explicit system message or empathy instruction is used in steering, separating activation-intervention effects from instruction effects.

\begin{table*}[t]
\centering
\small
\setlength{\tabcolsep}{5pt}
\begin{tabular}{p{0.24\textwidth}p{0.69\textwidth}}
\toprule
\textbf{Setting} & \textbf{Template} \\
\midrule
Teacher-forced extraction & \texttt{System: You are a helpful assistant.} \newline \texttt{User: The following is a post from someone seeking emotional support.} \newline \texttt{User: Seeker post: \{seeker\_post\}} \newline \texttt{Assistant: \{response\_post\}} \\
\midrule
Steering generation & \texttt{Seeker post: \{seeker\_post\}} \newline \texttt{Response:} \\
\midrule
OOD transfer on EmpatheticDialogues & \texttt{Situation: \{situation\}} \newline \texttt{Response:} \\
\midrule
Persona-conditioned generation & \texttt{You are a \{persona\}.} \newline \texttt{Seeker post: \{seeker\_post\}} \newline \texttt{Response:} \\
\bottomrule
\end{tabular}
\caption{Prompt templates used in the paper. The main text refers to this appendix rather than reproducing prompt strings inline.}
\label{tab:prompt-templates}
\end{table*}

\paragraph{Complete generation example.} The following is a Mistral-7B response under IP steering at layer 15 and $\alpha=+1$; its IP label changes from 0 in the corresponding unsteered generation to 2.
\begin{center}
\setlength{\fboxsep}{6pt}
\fbox{%
\begin{minipage}{0.92\columnwidth}
\small
\textbf{Input}\par
\texttt{Seeker post:} Today as I was walking home I stopped to lay on the grass for a while. It was nice. Just a reminder for everyone to get away from the electronics and just enjoy nature once in a while. \texttt{Response:}

\medskip
\textbf{Output}\par
I completely agree with you. In our fast-paced lives, we often forget to take a break and enjoy nature's calming effects. Taking time out to sit under a tree or lie down on the grass can be very rejuvenating. Nature has a way of reminding us that there is beauty in simplicity and it's important to appreciate it whenever possible. Thank you for sharing your experience!
\end{minipage}%
}
\end{center}

\subsection{GPT-5-mini Judge Prompts}
\label{sec:appendix-gpt-judge-prompts}

The cross-evaluator experiments in Section~\ref{sec:results-aux} use the OpenAI Batch API with \texttt{gpt-5-mini}. Each request contains a single text input and no separate system message. We use a separate rubric prompt for each EPITOME mechanism and ask the judge to output only the corresponding 0/1/2 label. The ER/IP/EX scoring instructions are taken from the original EPITOME annotation guidelines~\citep{sharma2020epitome}. Table~\ref{tab:gpt-judge-prompts} reproduces the prompt templates used in the experiment, reformatted only for compact presentation.

\begin{table*}[t]
\centering
\scriptsize
\setlength{\tabcolsep}{5pt}
\renewcommand{\arraystretch}{1.15}
\begin{tabular}{p{0.13\textwidth}p{0.80\textwidth}}
\toprule
\multicolumn{2}{c}{\textbf{GPT-5-mini Judge Prompt Templates}} \\
\midrule
\textbf{Common wrapper}
&
\textbf{Role:} You are an expert annotator for empathic responses. \newline
\textbf{Task:} Assess the candidate response. \newline
\textbf{Important:} Do not reward length by itself. Judge only the candidate response against the seeker post. \newline
\textbf{Input fields:} [Seeker Post] \texttt{\{seeker\_post\}}; [Candidate Response] \texttt{\{response\}}. \\
\midrule
\textbf{ER}
&
\textbf{Mechanism:} ER (Emotional Reactions). \newline
\textbf{Question:} Does the response express or allude to warmth, compassion, concern, or similar feelings of the responder toward the seeker? \newline
\textbf{Labels:} 0: No. \quad 1: Yes, alluded, but feelings are not explicitly expressed. \quad 2: Yes, explicit mention of such feelings. \newline
\textbf{Output exactly:} \texttt{ER: <0|1|2>} \\
\midrule
\textbf{IP}
&
\textbf{Mechanism:} IP (Interpretations). \newline
\textbf{Question:} Does the response communicate an understanding of the seeker's experiences and feelings? \newline
\textbf{Labels:} 0: No. \quad 1: Yes, communicates understanding. \quad 2: Yes, and includes one or more stronger interpretive behaviors, such as conjecture/speculation about the seeker's experiences and/or feelings; reflecting back on similar experiences of self/others; describing similar experiences of self/others; or paraphrasing the seeker's experiences and/or feelings. \newline
\textbf{Output exactly:} \texttt{IP: <0|1|2>} \\
\midrule
\textbf{EX}
&
\textbf{Mechanism:} EX (Explorations). \newline
\textbf{Question:} Does the response attempt to explore the seeker's experiences and feelings? \newline
\textbf{Labels:} 0: No. \quad 1: Yes, but exploration is generic. \quad 2: Yes, and exploration is specific. \newline
\textbf{Output exactly:} \texttt{EX: <0|1|2>} \\
\bottomrule
\end{tabular}
\caption{Prompt templates used for GPT-5-mini judging in the cross-evaluator experiment. The ER/IP/EX scoring instructions are taken from the original EPITOME annotation guidelines~\citep{sharma2020epitome}. Each batch request instantiates one mechanism-specific row together with the common wrapper and contains no separate system message.}
\label{tab:gpt-judge-prompts}
\end{table*}

\section{Prompt Baseline Details}
\label{sec:appendix-prompt-baseline}

This appendix supports Section~\ref{sec:results-prompt-baseline}. Table~\ref{tab:prompt-baseline-instructions} lists the exact instruction string used for each of the four prompt conditions. Each instruction is included in the user content before applying each model's chat template. P0 omits the instruction and therefore has the same user content as the unsteered steering prompt, but the serialized inputs are not token-identical because steering uses raw text. Table~\ref{tab:prompt-baseline-full} reports per-condition mean ER, IP, EX, and response token count across all three models and four prompt conditions.

\begin{table*}[t]
\centering
\small
\setlength{\tabcolsep}{5pt}
\begin{tabular}{p{0.20\textwidth}p{0.73\textwidth}}
\toprule
\textbf{Condition} & \textbf{Instruction string} \\
\midrule
P0 no\_instruction    & \textit{(no instruction)} \\
\midrule
P1 generic            & \texttt{Respond empathetically to the seeker.} \\
\midrule
P2 ER-targeted        & \texttt{Acknowledge the seeker's feelings warmly and compassionately.} \\
\midrule
P3 IP+EX-targeted     & \texttt{Reflect the seeker's situation, show understanding, and ask one open-ended follow-up question.} \\
\bottomrule
\end{tabular}
\caption{Instruction strings used in the prompt baseline (Section~\ref{sec:results-prompt-baseline}). Each instruction is included in the user content before chat-template serialization; P0 omits the instruction.}
\label{tab:prompt-baseline-instructions}
\end{table*}

\begin{table*}[t]
\centering
\small
\setlength{\tabcolsep}{4pt}
\begin{tabular}{ll|rrr|r}
\toprule
model & condition & ER mean & IP mean & EX mean & resp.\ tokens \\
\midrule
\multirow{4}{*}{Llama-3.1}
 & P0 no\_instruction    & 0.940 & 0.410 & 0.120 & 123.7 \\
 & P1 generic            & 1.240 & 0.550 & 0.050 & 126.7 \\
 & P2 ER-targeted        & 1.390 & 0.320 & 0.050 & 126.0 \\
 & P3 IP+EX-targeted     & 1.015 & 0.390 & 1.220 &  93.6 \\
\midrule
\multirow{4}{*}{Qwen2.5}
 & P0 no\_instruction    & 1.365 & 0.160 & 0.040 & 127.3 \\
 & P1 generic            & 1.585 & 0.610 & 0.020 & 127.1 \\
 & P2 ER-targeted        & 1.555 & 0.440 & 0.040 & 124.3 \\
 & P3 IP+EX-targeted     & 0.895 & 0.130 & 1.700 &  68.7 \\
\midrule
\multirow{4}{*}{Mistral-7B}
 & P0 no\_instruction    & 1.470 & 0.060 & 0.000 & 126.6 \\
 & P1 generic            & 1.790 & 0.200 & 0.000 & 126.1 \\
 & P2 ER-targeted        & 1.600 & 0.190 & 0.000 & 126.3 \\
 & P3 IP+EX-targeted     & 1.335 & 0.160 & 1.510 &  86.5 \\
\bottomrule
\end{tabular}
\caption{Per-condition mean ER, IP, EX, and response token count for the prompt baseline ($n=200$ per cell). P3 raises EX substantially on all three models but also reduces response length, and on Qwen and Mistral lowers ER.}
\label{tab:prompt-baseline-full}
\end{table*}

\section{Persona-Conditioned Steering Table}
\label{sec:appendix-persona-steer}

Table~\ref{tab:persona-steer} reports the per-$\alpha$ ER and IP means for the three race-related personas (\textit{person}, \textit{white person}, \textit{black person}) at layer 15 under $v_{15}^{\text{ER}}$ steering, summarized in Section~\ref{sec:results-projection}.

\begin{table}[h]
\centering
\small
\setlength{\tabcolsep}{4pt}
\begin{tabular}{lrrr|rrr}
\toprule
& \multicolumn{3}{c|}{ER} & \multicolumn{3}{c}{IP} \\
persona & $-$1 & 0 & $+$1 & $-$1 & 0 & $+$1 \\
\midrule
person       & 1.02 & 1.14 & 1.26 & 0.61 & 0.70 & 0.80 \\
\textit{white person} & 0.95 & 1.11 & 1.28 & 0.60 & 0.62 & 0.71 \\
\textit{black person} & 0.90 & 1.03 & 1.08 & 0.94 & 0.99 & 0.91 \\
\bottomrule
\end{tabular}
\caption{Race-conditioned ER and IP under $v_{15}^{\text{ER}}$ steering at $\alpha \in \{-1, 0, +1\}$ ($n=200$ each).}
\label{tab:persona-steer}
\end{table}

\section{Full Projection-Cross Table}
\label{sec:appendix-projection-full}

Table~\ref{tab:projection-cross-full} reports the full numeric source for Figure~\ref{fig:persona-projection}, with all persona categories included. It also reports IP-label and IP-projection shifts used in the discussion.

\begin{table*}[h]
\centering
\scriptsize
\setlength{\tabcolsep}{2pt}
\begin{tabular}{ll|rrrr|rrrr|rrrr}
\toprule
& & \multicolumn{4}{c|}{\textbf{Llama}} & \multicolumn{4}{c|}{\textbf{Qwen}} & \multicolumn{4}{c}{\textbf{Mistral}} \\
category & persona & $\Delta$ER & $\Delta p$ER & $\Delta$IP & $\Delta p$IP & $\Delta$ER & $\Delta p$ER & $\Delta$IP & $\Delta p$IP & $\Delta$ER & $\Delta p$ER & $\Delta$IP & $\Delta p$IP \\
\midrule
baseline & person & --- & --- & --- & --- & --- & --- & --- & --- & --- & --- & --- & --- \\
\midrule
\multirow{2}{*}{personality}
 & \textit{empathetic person} & $+$0.175 & $+$0.013 & $-$0.250 & $-$0.013 & $+$0.095 & $-$0.007 & $+$0.170 & $+$0.016 & $+$0.155 & $+$0.008 & $+$0.050 & $-$0.002 \\
 & \textit{cynical person}    & $-$0.550 & $-$0.055 & $-$0.330 & $+$0.025 & $-$0.735 & $-$0.035 & $+$0.060 & $+$0.039 & $-$0.385 & $-$0.035 & $+$0.090 & $+$0.017 \\
\midrule
\multirow{2}{*}{gender}
 & \textit{female} & $+$0.030 & $+$0.007 & $+$0.070 & $-$0.006 & $-$0.055 & $-$0.003 & $+$0.010 & $+$0.002 & $-$0.030 & $-$0.002 & $+$0.040 & $-$0.006 \\
 & \textit{male}   & $-$0.070 & $+$0.001 & $+$0.020 & $-$0.002 & $-$0.280 & $-$0.006 & $-$0.060 & $-$0.001 & $-$0.165 & $-$0.015 & $+$0.030 & $-$0.002 \\
\midrule
\multirow{2}{*}{race}
 & \textit{white person} & $-$0.040 & $+$0.016 & $-$0.050 & $-$0.019 & $-$0.110 & $+$0.003 & $-$0.040 & $-$0.001 & $-$0.130 & $-$0.015 & $+$0.070 & $-$0.001 \\
 & \textit{black person} & $-$0.105 & $+$0.011 & $+$0.480 & $+$0.007 & $-$0.100 & $+$0.008 & $-$0.060 & $-$0.001 & $-$0.130 & $-$0.011 & $+$0.050 & $-$0.001 \\
\midrule
\multirow{2}{*}{occupation}
 & \textit{psychotherapist} & $-$0.080 & $-$0.009 & $-$0.270 & $-$0.029 & $-$0.035 & $-$0.025 & $+$0.200 & $+$0.005 & $-$0.120 & $-$0.021 & $+$0.040 & $-$0.006 \\
 & \textit{engineer}        & $-$0.175 & $+$0.003 & $+$0.210 & $-$0.024 & $-$0.270 & $-$0.005 & $-$0.090 & $-$0.005 & $-$0.125 & $-$0.006 & $-$0.010 & $-$0.003 \\
\midrule
\multirow{2}{*}{political}
 & \textit{Democratic supporter} & $+$0.020 & $+$0.032 & $-$0.240 & $-$0.027 & $-$0.175 & $-$0.005 & $+$0.030 & $+$0.005 & $-$0.045 & $+$0.002 & $-$0.030 & $-$0.003 \\
 & \textit{Republican supporter} & $-$0.035 & $+$0.022 & $-$0.350 & $-$0.032 & $-$0.290 & $-$0.004 & $-$0.100 & $+$0.003 & $-$0.165 & $-$0.012 & $+$0.020 & $-$0.004 \\
\bottomrule
\end{tabular}
\caption{Full cross-model persona effects on ER/IP labels and on $v_{\text{ER}}$/$v_{\text{IP}}$ cosine projections at layer 15 ($n=200$ per persona), grouped by category. The ER columns are visualized in Figure~\ref{fig:persona-projection}.}
\label{tab:projection-cross-full}
\end{table*}

\section{Persona Shift Subspace Decomposition}
\label{sec:appendix-persona-subspace}

\begin{figure*}[h]
\centering
\includegraphics[width=\textwidth]{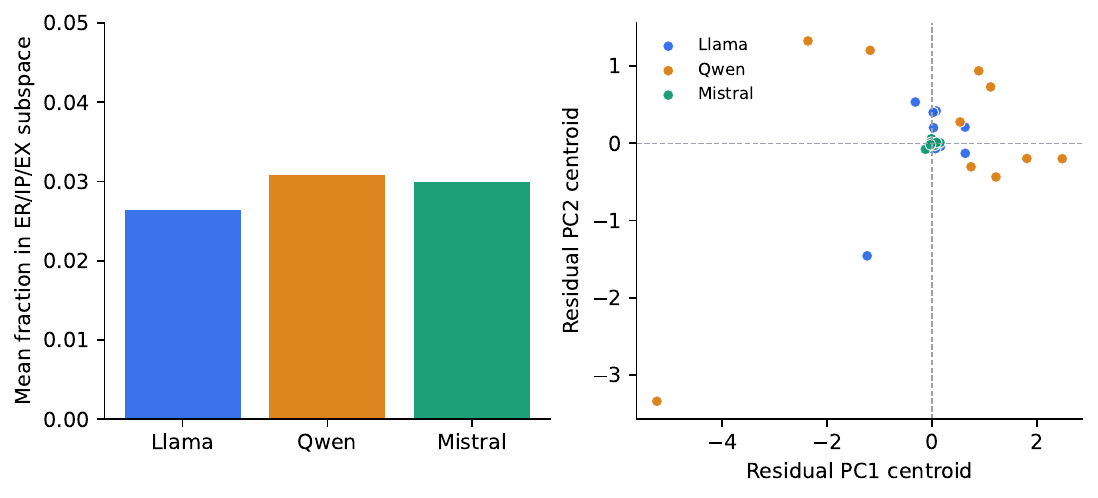}
\caption{Persona-shift decomposition. Left: mean persona-shift squared-norm fraction captured by the ER/IP/EX mechanism subspace. Right: persona centroids in the first two residual-PC coordinates. Residual PCs are computed after removing the ER/IP/EX mechanism-subspace component.}
\label{fig:persona-shift-subspace}
\end{figure*}

To test whether the persona result is only a failure of the ER axis or a broader mismatch with the recovered mechanism basis, we compute paired activation shifts $d_{p,i}=h_{p,i}-h_{\textit{person},i}$ for each non-baseline persona $p$ and seeker $i$ at layer 15. We then project $d_{p,i}$ onto an orthonormal basis for $\mathrm{span}(v_{\text{ER}},v_{\text{IP}},v_{\text{EX}})$ and compute the squared-norm fraction captured by this mechanism subspace. For persona-mean shifts, the reported fraction is $f_p=\|Q^\top \bar d_p\|^2/\|\bar d_p\|^2$, where $Q$ is an orthonormal basis for the ER/IP/EX span and $\bar d_p$ is the mean paired shift for persona $p$. Table~\ref{tab:persona-subspace-summary} reports both per-sample and persona-mean summaries. Figure~\ref{fig:persona-shift-subspace} visualizes the same decomposition and the first two residual-PC centroids. The PCA is descriptive: it shows structured residual variation, not a causal persona mechanism.

\paragraph{Scale reference.}
The mechanism subspace is 3-dimensional, embedded in the model residual stream ($d=4096$ for Llama and Mistral; $d=3584$ for Qwen). A uniformly random 3-D subspace would capture $3/d$ of the squared norm of an isotropic vector in expectation, below $0.1\%$ for all three models. The observed $2.6$--$3.1\%$ persona-mean fractions are therefore above this random-subspace scale reference, but still leave most persona-induced squared displacement outside the recovered mechanism basis. This comparison is only a scale reference, not a significance test: hidden states and persona shifts are not isotropic random vectors.

\begin{table}[h]
\centering
\scriptsize
\setlength{\tabcolsep}{8pt}
\begin{tabular}{lrrrr}
\toprule
Model & Sample & Mean & Resid. & Top-5 PCs \\
\midrule
Llama   & $1.85$ & $2.64$ & $97.36$ & $27.49$ \\
Qwen    & $2.63$ & $3.08$ & $96.92$ & $29.60$ \\
Mistral & $1.80$ & $3.00$ & $97.00$ & $29.86$ \\
\bottomrule
\end{tabular}
\caption{Persona-shift decomposition at layer 15. Columns report the percentage of squared shift norm captured by the orthonormalized ER/IP/EX mechanism subspace, the residual percentage for persona-mean shifts, and the cumulative percentage of residual variance explained by the top five residual PCs.}
\label{tab:persona-subspace-summary}
\end{table}

\end{document}